%% file: main.tex
\documentclass[letterpaper]{article} 
\usepackage[preprint]{aaai2027}
\usepackage[hyphens]{url}  
\usepackage{graphicx} 
\usepackage{natbib}  
\usepackage{caption} 
\usepackage{algorithm}
\usepackage{algorithmic}

\usepackage{newfloat}
\usepackage{listings}
\DeclareCaptionStyle{ruled}{labelfont=normalfont,labelsep=colon,strut=off} 
\floatstyle{ruled}
\newfloat{listing}{tb}{lst}{}
\floatname{listing}{Listing}

\usepackage{booktabs}
\usepackage{multirow}
\usepackage{xspace}
\usepackage{amsmath}
\usepackage{amssymb}
\newcommand{\cmark}{\checkmark}

\newcommand{\eg}{\emph{e.g.}\xspace}
\newcommand{\ie}{\emph{i.e.}\xspace}

\title{Hyper-FEOD: Sparse Hypergraph-Enhanced Frame-Event Object Detection with \\ Fine-Grained MoE}

\title{Hyper-FEOD: Sparse Hypergraph-Enhanced Frame-Event Object Detection with \\ Fine-Grained MoE}
\author {
    Wei Bao\textsuperscript{\rm 1},
    Yuehan Wan\textsuperscript{\rm 1},
    Tianhang Zhou\textsuperscript{\rm 2},
    Siqi Li\textsuperscript{\rm 1}\corresponding
}
\affiliations {
    \textsuperscript{\rm 1}BNRist, THUIBCS, BLBCI, School of Software, Tsinghua University \\
    \textsuperscript{\rm 2}State Key Laboratory of Heavy Oil Processing, China University of Petroleum (Beijing), Beijing, 102249, China\\
    baoweivvv@gmail.com, wangyuehan@stu.xjtu.edu.cn, zhouth@cup.edu.cn, lisiqi19971013@gmail.com
}

\begin{document}

\maketitle

\input{0_abstract}

\input{1_intro}

\input{2_related}

\input{3_method}

\input{4_exp}

\input{5_conclusion}

\bibliography{aaai2027}


\end{document}

%% file: 0_abstract.tex
\begin{abstract}
The integration of frame-based RGB cameras with event streams constitutes a promising paradigm for robust object detection under challenging dynamic conditions. Nevertheless, effectively modeling intricate multi-modal interactions and reconciling the semantic heterogeneity between RGB and event data remain formidable challenges for high-precision detection. In this paper, we present Hyper-FEOD, a novel high-performance detection framework that synergistically strengthens cross-modal representation learning through two core hypergraph-driven components. Specifically, we first design a Sparse Hypergraph-enhanced Cross-Modal Fusion (SHCF) module that exploits event activity cues to identify motion-critical sparse tokens and performs high-order relational reasoning through hypergraph modeling. This design effectively captures intricate high-order dependencies and rich contextual correlations across modalities. Second, we develop a Fine-Grained Mixture-of-Experts (FG-MoE) module tailored to handle the heterogeneous semantic demands arising from distinct visual regions. By deploying specialized hypergraph experts with varying hyperedge connectivity pattern and incorporating a spatial gating mechanism, FG-MoE adaptively routes features to enable precise enhancement at target regions. Coupled with an auxiliary router loss, the proposed framework ensures stable end-to-end training and optimal feature refinement. Comprehensive experiments conducted on widely-adopted RGB-Event benchmarks show that Hyper-FEOD delivers superior detection performance and consistently outperforms existing state-of-the-art approaches by a notable margin.
\end{abstract}

%% file: 1_intro.tex
\section{Introduction}
\label{sec:intro}
Object detection serves as a fundamental and pivotal perception task in computer vision~\cite{liu2020deep,oksuz2020imbalance}. Conventional frame-based RGB cameras, although providing rich color, texture, and semantic information, are constrained by fixed exposure time and limited dynamic range, resulting in severe performance degradation under challenging conditions such as high-speed motion, low illumination, and abrupt lighting changes~\cite{perot2020learning}. In contrast, emerging bio-inspired event cameras (\eg, DAVIS) operate in an asynchronous manner and only transmit local pixel-level intensity changes~\cite{gallego2022event}. Endowed with microsecond-level temporal resolution and high dynamic range, event cameras effectively alleviate motion blur and excel in extreme scenarios, offering a promising complement to the inherent shortcomings of RGB cameras~\cite{gehrig2023recurrent}. However, event streams intrinsically suffer from data sparsity and the absence of dense appearance cues, particularly for static objects and fine textures.
\begin{figure}[t]
    \centering\includegraphics[width = 1\linewidth]{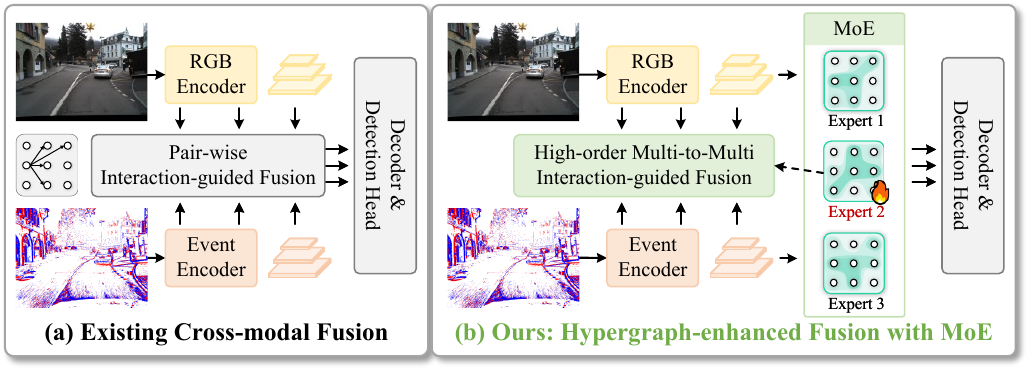}
    \caption{Comparison with existing frame-event fusion paradigms. (a) Prior methods adopt pair-wise interaction with a fixed topology. (b) Our Hyper-FEOD performs high-order multi-to-multi hypergraph interaction, coupled with a Mixture-of-Experts (MoE) for content-adaptive fusion.}
    \label{fig:intro}
    \vspace{-0.3cm}
\end{figure}

The intrinsic complementarity between frame and event modalities has spurred rapid development of Frame-Event Object Detection (FEOD) methods~\cite{gehrig2024low,zhou2022rgb,li2023sodformer,zhang2024frequency}. DAGr~\cite{gehrig2024low} exploits the high temporal resolution of events together with the rich semantics of RGB images to achieve efficient and high-frequency detection. RENet~\cite{zhou2022rgb} designs a bi-directional feature calibration scheme to attentively form the shared cross-modal representation. SODFormer~\cite{li2023sodformer} integrates events and frames via a spatiotemporal transformer for continuous streaming detection. FAOD~\cite{zhang2024frequency} further addresses the frequency mismatch between low-rate frames and high-rate events through attention-based cross-modal alignment. Despite substantial progress, the detection performance of existing FEOD methods remains suboptimal in complex scenarios, which we attribute to two under-explored limitations. On the one hand, prevailing fusion designs rely on dense cross-attention or graph construction and lack the capacity to model global multi-to-multi high-order correlations under the intrinsic sparsity of event data, as illustrated in Fig.~\ref{fig:intro}(a). Treating all spatial positions as equally important is not only computationally wasteful but also dilutes motion-critical cues that are naturally concentrated around brightness changes. On the other hand, existing frameworks typically adopt a uniform interaction topology across all samples, thereby overlooking the heterogeneous semantic demands arising from distinct visual regions and dynamic scenes. Since event density, object scale, ego-motion, and background clutter vary drastically across samples, fixed hyperedge connectivity pattern inevitably leads to insufficient exploitation of cross-modal complementarity.

To address the aforementioned limitations, this work presents a novel Sparse \textbf{Hyper}graph-enhanced \textbf{F}rame-\textbf{E}vent \textbf{O}bject \textbf{D}etection framework with Fine-Grained MoE (\textbf{Hyper-FEOD}), which advances cross-modal fusion from two perspectives: sparse high-order interaction and content-adaptive specialization, as depicted in Fig.~\ref{fig:intro}(b). For high-order cross-modal interaction, we introduce a Sparse Hypergraph-enhanced Cross-Modal Fusion (SHCF) module, which converts the intrinsic sparsity of event streams into a per-sample activity prior and adaptively selects motion-critical target nodes while retaining dense multimodal context. Explicit binary hyperedges are then built only for the selected targets to aggregate features from both modalities and disseminate them back to the corresponding spatial grid, enabling efficient multi-to-multi high-order relational reasoning across modalities. For content-adaptive specialization, we further develop a Fine-Grained Mixture-of-Experts (FG-MoE) module tailored to the heterogeneous semantic demands of different scenes. FG-MoE instantiates a pool of structurally diverse SHCF experts with distinct neighborhood scales and retention budgets. Coupled with load-balancing regularization and router z-loss, FG-MoE enables stable end-to-end training and precisely allocates high-order computation to regions without executing the entire expert pool. By synergistically integrating these two components, Hyper-FEOD fully exploits the complementarity of frame and event modalities and consistently outperforms existing state-of-the-art detectors by a notable margin. The main contributions of our paper are summarized as follows:
\begin{itemize}
    \item We propose \textbf{Hyper-FEOD}, a novel frame-event object detection framework that advances cross-modal fusion through sparse high-order interaction and content-adaptive specialization, effectively exploiting the complementarity between frame and event modalities.
    \item We design a Sparse Hypergraph-enhanced Cross-Modal Fusion (SHCF) module, which leverages intrinsic event sparsity as a per-sample activity prior to construct hyperedges only on motion-critical nodes, enabling efficient high-order cross-modal reasoning while avoiding the redundancy of dense pairwise interaction.
    \item A Fine-Grained Mixture-of-Experts (FG-MoE) module is proposed to deploy structurally diverse hypergraph experts with heterogeneous connectivity patterns, and adaptively route features via sample-wise gating to accommodate diverse scenes with controllable cost.
    \item Extensive experiments on two widely used RGB-Event object detection benchmarks demonstrate that Hyper-FEOD achieves superior performance and establishes new state-of-the-art (SOTA) results.
\end{itemize}

%% file: 2_related.tex
\section{Related Work}
\label{sec:related_work}

\subsection{Erame-Event Object Detection}
Event cameras encode brightness changes rather than conventional intensity frames, offering high temporal resolution, high dynamic range, and reduced motion blur~\cite{gallego2022event}. These properties have supported
large-scale driving datasets such as DSEC~\cite{gehrig2021dsec} and specialized
event-based detectors. Recurrent Vision Transformers (RVT) combine local and
dilated attention with recurrent temporal aggregation for event-only object
detection~\cite{gehrig2023recurrent}. Such methods exploit event dynamics well, but
the absence of absolute intensity makes static appearance and fine texture
difficult to recover from events.

Frame-Event detection instead combines complementary appearance and motion
cues. RENet aggregates event representations over multiple temporal windows
and performs bidirectional feature calibration for moving-object detection
~\cite{zhou2022rgb}. SODFormer uses spatiotemporal transformers and
asynchronous attention to support streaming detection from events and frames
~\cite{li2023sodformer}. FAOD explicitly addresses the frequency mismatch
between low-rate RGB frames and high-rate events through alignment and
time-shift training~\cite{zhang2024frequency}. Recently, PEOD provides a
pixel-aligned, high-resolution benchmark and reports that multimodal fusion
still has substantial room for improvement under severe illumination
degradation~\cite{cui2025peod}. Different from these methods, Hyper-FEOD  treats event activity as a
computational prior, selects a different sparse target set for each sample, and uses dense RGB and event features only as contextual nodes around those
targets. This preserves multimodal context while avoiding uniform high-order processing over inactive regions.
\begin{figure*}[t]
    \centering\includegraphics[width = 1\linewidth]{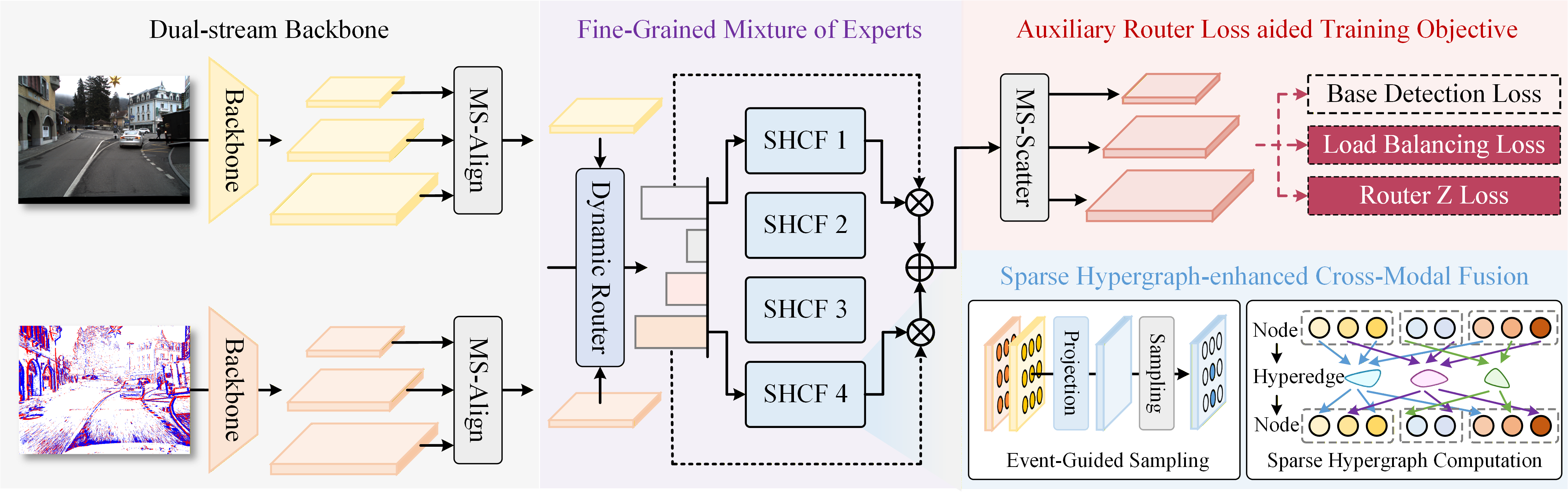}
    \caption{Overall architecture of the proposed Hyper-FEOD. A dual-stream backbone first extracts multi-scale features from RGB frames and event streams, which are aligned into middle level via MS-Align. A Fine-Grained Mixture-of-Experts (FG-MoE) then routes the multi-modal aligned features through a dynamic router to a pool of Sparse Hypergraph-enhanced Cross-Modal Fusion (SHCF) experts, each performing event-guided adaptive sampling and sparse hypergraph computation for high-order multi-to-multi cross-modal reasoning. The fused features are scattered back into multi-level features via MS-Scatter and supervised by a base detection loss together with auxiliary load-balancing and router z-losses.}
    \label{fig:overall}
    \vspace{-0.3cm}
\end{figure*}

\subsection{Visual Hypergraph Computation}
Hypergraph is the extension of graph. A hypergraph $\mathcal G$ can be defined by the hyperedge set $\mathcal E$ and the vertex set $\mathcal V$, \ie, $\mathcal G=\{\mathcal E, \mathcal V\}$. Different from traditional graphs, each hyperedge of a hypergraph can connect multiple vertices, and thus complex correlations between multiple vertices can be modeled. Hypergraph learning can facilitate the correlation-guided analysis by capturing complex and higher-order correlations in data based on hypergraph structures. Gao~\cite{hgnn} propose a higher-order message spatial propagation method between vertices, which further enhance the capability of hypergraph learning. Recently, several works~\cite{han2023vision,lei2025softhgnn,fixelle2025hypergraph} have applied hypergraph computation to computer vision. HyperYOLO~\cite{feng2024hyper} is the first work to apply hypergraph computation to object detection, proposing a semantic collecting and scattering framework and modeling hyperedges as $\epsilon$-ball centered at each feature point. YOLOv13~\cite{lei2025yolov13} advances hyperedge modeling by formulating them as global features, where each feature is generated via learnable prototypes combined with global offsets to facilitate hypergraph message propagation in a soft-connection manner. Notably, our method is the first work to apply hypergraph computation to FEOD, designing hyper attention to enhance intra- and inter-modality high-order information interaction and enable efficient cross-modal fusion.

\subsection{Mixture of Experts}
Mixture-of-experts models increase representational capacity through
input-dependent conditional computation. Sparsely-gated MoE introduced learned
routing with an auxiliary balancing objective~\cite{shazeer2017outrageously}, and
Switch Transformers simplified sparse routing while highlighting optimization
and numerical-stability issues~\cite{fedus2022switch}. In computer vision,
V-MoE routes image tokens to sparse feed-forward experts and demonstrates
adaptive computation in vision transformers~\cite{riquelme2021scaling}. ST-MoE
further studies stable sparse-expert training and introduces the router z-loss
used in our formulation~\cite{zoph2022st}. FG-MoE differs from conventional feed-forward expert pools in both expert
structure and routing granularity. Each expert is an SHCF operator with a
particular node-budget prior and neighborhood scale, rather than an
architecturally identical multilayer perceptron. Routing is performed once per
sample, not independently for every pixel or token, which matches the
per-sample sparse-node construction and limits dispatch overhead. A shared
expert and load balancing provide more
stable common path while allowing high-order interaction.

%% file: 3_method.tex
\section{Method}
In this section, we first introduce the overall framework of our proposed Hyper-FEOD. Then, we introduce the sparse hypergraph-enhanced cross-modal fusion, fine-grained mixture of experts and the overall training objective. 
\label{sec:method}
\subsection{Overview}
\label{subsec:overview}
As illustrated in Fig., Hyper-FEOD employs a dual-stream backbone to derive multiscale feature pyramids $\{\mathbf{F}^{m}_{3},\mathbf{F}^{m}_{4},\mathbf{F}^{m}_{5}\}$ from both RGB and event modalities ($m\in\{r,e\}$). To facilitate high-order interaction while mitigating scale redundancy, the multiscale features of each stream are aligned and projected into a shared intermediate representation space centered at middle level:
\begin{equation}
\begin{split}
    \mathbf{Z}^{m} ={} & \psi^{m}_{4}\left(\mathbf{F}^{m}_{4}\right) + \beta^{m}_{3}\,\mathcal{A}_{3\rightarrow4}\left(\psi^{m}_{3}(\mathbf{F}^{m}_{3})\right) \\
    & + \beta^{m}_{5}\,\mathcal{A}_{5\rightarrow4}\left(\psi^{m}_{5}(\mathbf{F}^{m}_{5})\right), \quad m\in\{r,e\},
\end{split}
\end{equation}
where $\psi^m_s$ denotes a $1\times1$ projection, $\mathcal{A}_{s\rightarrow4}$ is a spatial alignment operator (downsampling or upsampling), and $\beta_s^m$ are learnable channel-wise coefficients initialized to zero.

The resultant intermediate features $\mathbf{Z}^{r}$ and $\mathbf{Z}^{e}$ are subsequently forwarded to the Sparse Hypergraph-enhanced Cross-Modal Fusion (SHCF) module to establish high-order contextual correlations and achieve multimodal integration. The fused representation $\mathbf{Z}^{fuse}$ subsequently injected into the original multiscale RGB features so as to preserve the rich appearance and structural priors:
\begin{equation}
\begin{aligned}
    \mathbf{F}_3^{fuse} &= \mathbf{F}^{r}_{3} + \mathbf{\Gamma}_{3} \odot \mathcal{A}_{4\rightarrow3}(\mathbf{Z}^{fuse}) \\
    \mathbf{F}_4^{fuse} &= \mathbf{F}^{r}_{4} + \mathbf{\Gamma}_{4} \odot \mathbf{Z}^{fuse} \\
    \mathbf{F}_5^{fuse} &= \mathbf{F}^{r}_{5} + \mathbf{\Gamma}_{5} \odot \mathcal{A}_{4\rightarrow5}(\mathbf{Z}^{fuse})
\end{aligned}
\label{eq:residual_injection}
\end{equation}
where $\mathbf{\Gamma}_s$ is a channel-wise gate. To further boost the model's capacity and adaptivity, Hyper-FEOD integrates a Fine-Grained Mixture of Experts (MoE) mechanism within the cross-modal fusion stage. The MoE architecture dynamically dispatches sample representations to a specialized pool of experts characterized by diverse neighborhood ranges and sampling budgets. Moreover, inspired by sparse MoE routing paradigms, a router load-balancing loss is incorporated to alleviate expert collapse and stabilize the routing behavior.

\subsection{Sparse Hypergraph-enhanced Cross-Modal Fusion}
\label{subsec:sparse_hypergraph}
To efficiently capture high-order contextual correlations, SHCF operates in two main stages: (1) Event-Guided Adaptive Node Sampling, which extracts high-value target nodes to reduce spatial redundancy, and (2) Sparse Hypergraph Computation, which models high-order relation across modalities.

\subsubsection{Event-Guided Adaptive Node Sampling}
\label{subsubsec:node_sampling}
Given aligned RGB and event features $\mathbf{Z}^{r}$ and $\mathbf{Z}^{e}$, we first form a joint representation $\mathbf{Z} = \phi_{\mathrm{joint}}([\mathbf{Z}^{r}; \mathbf{Z}^{e}]) \in \mathbb{R}^{N \times C}$, refined by a multi-branch dilated convolution unit with CBAM attention. To avoid dense spatial graph overhead, we dynamically select a subset of target nodes guided by event activity. For a sample $b$ and expert $q$, an event activity map $\mathbf{A}_{b} = \operatorname{MaxPool}_{3\times3}(\frac{1}{C}\sum_{c=1}^{C}|\mathbf{Z}_{b,c}|)$ is computed and log-normalized into $\widehat{\mathbf{A}}_{b}$. A unified sampling score map $\mathbf{S}_{bq} \in \mathbb{R}^N$ and sample-adaptive retention budget $K_{bq}$ are formulated as:
\begin{equation}
\begin{aligned}
    \mathbf{S}_{bq} &= \phi^{q}_e(\mathbf{Z}_{e}) + \lambda_a\widehat{\mathbf{A}}_{b} + \lambda_r\phi^{q}_r(\mathbf{Z}_{r}) \\ \quad K_{bq} &= \operatorname{clip}\left(\left\lfloor \rho_{bq}N\right\rfloor, 1, N\right),
\end{aligned}
\label{eq:selection_score}
\end{equation}
where $\rho_{bq}$ dynamically scales with the scene event density $d_b$. Using the top-$K_{bq}$ indices $\mathcal{I}_{bq}$ derived from $\mathbf{S}_{bq}$, we gather the target nodes $\mathbf{X}^{bq}_t = \operatorname{Gather}(\mathbf{Z}_b, \mathcal{I}_{bq}) \in \mathbb{R}^{K_{bq}\times C}$. Meanwhile, all spatial tokens from both modalities are flattened to preserve complete global context $\mathbf{X}^b_c = [\operatorname{vec}(\mathbf{Z}^r_b); \operatorname{vec}(\mathbf{Z}^e_b)] \in \mathbb{R}^{2N\times C}$.

\subsubsection{Sparse Hypergraph Computation}
\label{subsubsec:hypergraph_prop}
We construct a heterogeneous hypergraph $\mathcal{G} = (\mathcal{V}, \mathcal{E})$ to connect the downsampled target nodes with the full contextual nodes. The node set $\mathcal{V}$ consists of both target and contextual nodes, i.e., $\mathbf{X}^{bq}_{\mathrm{all}} = [\mathbf{X}^{bq}_t; \mathbf{X}^b_c] \in \mathbb{R}^{(K_{bq}+2N)\times C}$. Each selected target node $i \in \{1, \dots, K_{bq}\}$ defines a hyperedge $e_i \in \mathcal{E}$ that grouping its $k_q$-nearest neighbors within $\mathbf{X}^{bq}_{\mathrm{all}}$ under normalized cosine affinity $a^{bq}_{ij}$. The explicit binary incidence matrix $\mathbf{H}^{bq} \in \{0, 1\}^{(K_{bq}+2N) \times K_{bq}}$ is defined as:
\begin{equation}
    \mathbf{H}^{bq}_{v, e_i} = 
    \begin{cases} 
      1, & \text{if node } v \in \mathcal{N}_{k_q}(i) \text{ or } v = i, \\
      0, & \text{otherwise,}
    \end{cases}
    \label{eq:incidence_matrix}
\end{equation}
where $\mathcal{N}_{k_q}(i)$ represents the set of $k_q$-nearest contextual/target neighbors for hyperedge $e_i$.

High-order message passing proceeds along a dual-stage Node-to-Hyperedge-to-Node pathway:
\begin{equation}
\begin{aligned}
    \mathbf{Y}^{bq}_{e} &= (\mathbf{D}^{bq}_{e})^{-1}(\mathbf{H}^{bq})^\mathsf{T}\left(\mathbf{X}^{bq}_{\mathrm{all}}W_{n2e}\right) \\ \quad \mathbf{Y}^{bq}_{v} &= (\mathbf{D}^{bq}_{v})^{-1}\mathbf{H}^{bq}\left(\mathbf{Y}^{bq}_{e}W_{e2n}\right),
\end{aligned}
\end{equation}
where $\mathbf{D}^{bq}_{e} \in \mathbb{R}^{K_{bq} \times K_{bq}}$ and $\mathbf{D}^{bq}_{v} \in \mathbb{R}^{(K_{bq}+2N) \times (K_{bq}+2N)}$ are diagonal degree matrices for hyperedges and nodes, respectively, and $W_{n2e}, W_{e2n}$ are learnable projection weights. After layer normalization and residual connection $\widetilde{\mathbf{X}}^{bq}_{\mathrm{all}}=\operatorname{LN}(\mathbf{Y}^{bq}_v + \mathbf{X}^{bq}_{\mathrm{all}})$, the updated target node features are extracted and scatter-added back to their original spatial grid positions in $\mathbf{Z}_b$ to get fused feature $\mathbf{Z}^{fuse}$. This ensures high-order context propagation while preserving non-selected background features with minimal computational cost.

\subsection{Fine-Grained Mixture of Experts}
\label{subsec:moe}
To handle dramatic scene variations in object scale and dynamic event density, we embed a sample-wise Fine-Grained Mixture-of-Experts (MoE) architecture within the SHCF fusion stage. Instead of applying a uniform hypergraph topology across all samples, our framework employs a dynamic pool of specialized hypergraph experts $\{\mathcal{G}_q\}_{q=1}^E$. 

The experts in the pool are customized with distinct neighborhood search ranges $k_q \in \{2, 4, 8\}$ and retention ratio intervals $[\rho_{\min}^q, \rho_{\max}^q]$, enabling them to capture multi-scale contexts ranging from localized fine details to broad global structures. Given the concatenated features $[\mathbf{Z}^r_b; \mathbf{Z}^e_b]$ for sample $b$, the router predicts expert logits $\mathbf{g}_b \in \mathbb{R}^E$ and soft probability distribution $\mathbf{p}_b$:
\begin{equation}
    \mathbf{g}_b = \operatorname{GAP}\left(\phi_{\mathrm{route}}\left([\mathbf{Z}^r_b; \mathbf{Z}^e_b]\right)\right), \quad p_{bq} = \frac{\exp(g_{bq} / T_r)}{\sum_{\ell=1}^E \exp(g_{b\ell} / T_r)},
    \label{eq:router_logits}
\end{equation}
where $\operatorname{GAP}(\cdot)$ denotes global average pooling, $\phi_{\mathrm{route}}$ is a lightweight projection head, and $T_r$ is a temperature hyperparameter.

To perform sparse activation, the router selects the Top-$K_m$ experts with the highest probabilities, denoted as $\mathcal{S}_b = \operatorname{TopK}(\mathbf{p}_b, K_m)$. The gate values for the active experts are renormalized as:
\begin{equation}
    \widetilde{p}_{bq} = \frac{p_{bq}}{\sum_{\ell \in \mathcal{S}_b} p_{b\ell} + \epsilon}, \quad \forall q \in \mathcal{S}_b.
    \label{eq:renorm_gate}
\end{equation}
In addition to the dynamically routed experts, a lightweight shared expert $\mathcal{G}_{\mathrm{sh}}$ with a fixed global field remains continuously active to preserve foundational cross-modal alignment and maintain a stable primary execution path. The final fused representation $\mathbf{Z}^{fuse}_b$ for sample $b$ is computed by aggregating the shared path with the gated expert outputs:
\begin{equation}
    \mathbf{Z}^{fuse}_b = \sum_{q \in \mathcal{S}_b} \widetilde{p}_{bq} \mathcal{G}_{q}\left(\mathbf{Z}_b, \mathbf{Z}^r_b, \mathbf{Z}^e_b\right),
    \label{eq:moe_output}
\end{equation}
where $\mathcal{G}_{q}$ denotes the sparse hypergraph execution under expert $q$'s specific hyperparameters $(k_q, \rho_{bq})$.

During training, routing soft decisions allow backpropagation across all candidate expert paths to encourage diversity. During inference, we leverage conditional dispatch: each sample is routed exclusively to its assigned $K_m^{\mathrm{test}} = 2$ experts (out of $E=8$), skipping the evaluation of unselected experts entirely. This dynamic sparse execution substantially lowers FLOPs and latency while retaining high model capacity across diverse visual scenarios.

\subsection{Overall Loss}
\label{subsec:loss}
To train the Hyper-FEOD network end-to-end, we adopt a joint objective function combining the primary object detection loss $\mathcal{L}_{\mathrm{det}}$ with an auxiliary router loss $\mathcal{L}_{\mathrm{aux}}$:
\begin{equation}
    \mathcal{L} = \mathcal{L}_{\mathrm{det}} + \lambda_{\mathrm{moe}}\mathcal{L}_{\mathrm{aux}},
    \label{eq:total_loss}
\end{equation}
where $\mathcal{L}_{\mathrm{det}}$ integrates standard bounding box regression, classification, and distribution focal losses. To prevent expert collapse—where the router continuously selects a small subset of experts—we incorporate an auxiliary loss $\mathcal{L}_{\mathrm{aux}} = \lambda_{\mathrm{bal}}\mathcal{L}_{\mathrm{bal}} + \lambda_z \mathcal{L}_{z}$ to stabilize MoE optimization.

For a batch of $B$ samples, the mean gating probability $v_q$ (expert importance) and hard selection frequency $u_q$ (expert load) for expert $q$ are defined as:
\begin{equation}
    v_q = \frac{1}{B}\sum_{b=1}^B p_{bq}, \quad u_q = \frac{1}{B K_m}\sum_{b=1}^B \mathbb{I}\left[q \in \mathcal{S}_b\right],
    \label{eq:expert_load}
\end{equation}
where $\mathbb{I}[\cdot]$ is the indicator function. The load-balancing loss encourages uniform distribution across all $E$ experts:
\begin{equation}
    \mathcal{L}_{\mathrm{bal}} = E \sum_{q=1}^{E} v_q \cdot \operatorname{sg}(u_q),
    \label{eq:l_bal}
\end{equation}
where $\operatorname{sg}(\cdot)$ denotes the stop-gradient operator.

To prevent numerical instability caused by excessively large router logits $\mathbf{g}_b$, we penalize the exponentiated partition function via:
\begin{equation}
    \mathcal{L}_{z} = \frac{1}{B}\sum_{b=1}^{B}\left(\log \sum_{q=1}^{E}\exp(g_{bq})\right)^2.
    \label{eq:router_z_loss}
\end{equation}

%% file: 4_exp.tex
\section{Experiments}
\label{sec:experiments}
In this section, we first elaborate on the experimental settings, including datasets, implementation details, and metrics. Subsequently, quantitative comparisons between our proposed method and SOTA approaches are presented. Ablation experiments are performed to verify the effectiveness of each proposed module. Finally, visualization results are provided.
\subsection{Experimental Settings}
\paragraph{Datasets.} We conduct experiment on two widely used datasets: DSEC-Detection and PKU-DAVIS-SOD.

\textbf{DSEC-Detection} extends the DSEC driving dataset
\cite{gehrig2021dsec,gehrig2024low} with object-detection annotations.
The released benchmark contains 60 sequences, 70,379 annotated frames,
and 390,118 bounding boxes from eight traffic categories: car,
pedestrian, rider, motorcycle, bicycle, truck, bus, and train. It
combines a $640\times480$ event stream with temporally corresponding
frame-camera observations and includes challenging urban, nighttime,
and high-dynamic-range scenes.

\textbf{PKU-DAVIS-SOD} was introduced with SODFormer
\cite{li2023sodformer}. It contains 220 RGB-Event driving sequences
captured with a DAVIS346 sensor at a spatial resolution of
$346\times260$. The dataset provides approximately 276k labeled
timestamps and 1.0801M manually annotated boxes at 25\,Hz for three
classes (car, pedestrian, and two-wheeler), covering normal, low-light,
and motion-blur conditions.

\paragraph{Implementation Details.}
We instantiate Hyper-FEOD at the YOLO11s model scale with two
modality-specific streams. Frame-aligned event images and their paired
RGB observations are provided to the network through the six-channel
multimodal data pipeline. The SHCF/FG-MoE block contains eight
candidate experts and activates the top two routed experts at
inference. The auxiliary MoE-loss coefficient is set to
$\lambda_{\mathrm{moe}}=1.5$.

Training is scheduled for at most 100 epochs with automatic optimizer
selection, mixed-precision computation, a three-epoch warm-up, and
weight decay of $5\times10^{-4}$. We use random horizontal flipping,
translation, scaling, and mosaic augmentation, disabling mosaic during
the last ten epochs. On DSEC-Detection, inputs are resized to
$640\times640$ and the batch size is 8; training uses two NVIDIA RTX
3090 GPUs. On PKU-DAVIS-SOD, the corresponding settings are
$346\times346$, a batch size of 32, and two NVIDIA Tesla V100 32\,GB
GPUs.

\paragraph{Evaluation Metrics.}
Following the COCO protocol \cite{lin2014microsoft}, we report
$\mathrm{mAP}$ averaged over IoU thresholds from 0.50 to 0.95. Parameters and FLOPs are also provided.

\subsection{Comparison with State-of-the-Art Methods}

\input{tab1}
Table~\ref{tab:tab1} summarizes the quantitative comparison between our Hyper-FEOD and existing multimodal frame-event detection methods on two benchmarks. On the PKU-DAVIS-SOD dataset, Hyper-FEOD achieves 32.3 mAP and 58.5 AP$_{50}$, outperforming the previous state-of-the-art FAOD by 1.6 mAP and 1.0 AP$_{50}$ respectively, with a slight acceptable latency increase of 1.0 ms. For the more challenging DSEC-Detection driving benchmark, our method obtains a substantial performance boost: it reaches 52.5 mAP and 73.2 AP$_{50}$, surpassing FAOD by 10.0 mAP and 9.7 AP$_{50}$. Such significant gains demonstrate that the performance superiority of Hyper-FEOD does not merely stem from simple multimodal feature concatenation. Instead, the sparse hypergraph high-order reasoning and fine-grained adaptive MoE routing effectively allocate cross-modal interaction computation to motion-critical regions, fully exploiting the complementary appearance cues from RGB frames and motion information from event streams under complex night, high-dynamic-range and high-speed motion driving scenes. In terms of model scale, the total stored parameters of Hyper-FEOD are 19.42 M, while only 14.42 M parameters are activated during inference via Top-2 expert routing, which is comparable or even lighter than most competing fusion methods, verifying the efficiency of our sparse conditional computation paradigm.

\subsection{Ablation Studies}
We conduct comprehensive ablation experiments on DSEC-Detection to dissect the independent and joint contributions of each core component.

\paragraph{Effectiveness of SHCF and FG-MoE.}
\input{tab2}
Table \ref{tab:tab2} validates the complementary gains brought by the two core proposed modules. The baseline only achieves 50.9 mAP and 71.5 AP$_{50}$. Equipping the model with the standalone SHCF improves mAP by 0.7 points to 51.6, which proves that event-guided sparse high-order cross-modal relational reasoning can capture richer multi-modal contextual correlations than dense pairwise interaction. When replacing SHCF with ordinary convolution-based MoE (Row 3), the model gains 1.0 mAP to 51.9, demonstrating that conditional expert routing can adapt feature processing to heterogeneous scene semantics. Combining SHCF and FG-MoE together yields the optimal 52.5 mAP and 73.2 AP$_{50}$, with a total improvement of 1.6 mAP over the baseline. 

\paragraph{Breakdown of SHCF Sub-components.}
\input{tab3}
Table \ref{tab:tab3} decouples the two core stages of the SHCF module: event-guided sparse node sampling and hypergraph high-order message passing. Activating sparse sampling alone brings a 0.3 mAP improvement, as filtering motion-irrelevant background tokens reduces redundant spatial computation and suppresses noisy background interference. Applying hypergraph modeling over all spatial tokens without sparse sampling lifts mAP to 51.4, which verifies that multi-to-multi hyperedges can model complex high-order cross-modal dependencies ignored by pairwise graph fusion. Integrating sparse sampling and hypergraph computation together achieves the maximum 51.6 mAP, showing that event activity prior can effectively select critical motion nodes to construct lightweight hypergraphs, simultaneously retaining complete multimodal context and eliminating dense spatial computation overhead.

\paragraph{Impact of Experts Number.}
\input{tab4}
Table~\ref{tab:tab4} analyzes the performance variation with different numbers of hypergraph experts in the FG-MoE. It can be seen that expanding the expert pool from 4 to 8 continuously boosts detection accuracy. Although the total stored model parameters increase with more experts, the number of active parameters at inference remains tightly controlled by sparse routing, avoiding linear growth of computation cost. This result validates the core advantage of FG-MoE: we can learn a richer set of specialized hypergraph transformation patterns to adapt to diverse object scales and event density distributions, without executing the full expert pool for each input sample.

\paragraph{Top-$K$ Selection.}
\input{tab5}
Table \ref{tab:tab5} explores the accuracy-efficiency trade-off by adjusting the number of activated experts per samplel. When all 8 experts are fully activated (sparsity 0\%), the model achieves the lowest mAP of 52.0, as redundant expert computation introduces feature noise and increases optimization difficulty. Activating 6 or 4 experts moderately reduces model performance to 52.2 mAP. The optimal performance is achieved at Top-2 expert routing with 75\% expert sparsity, reaching 52.5 mAP and 73.2 AP$_{50}$. This demonstrates that only a small subset of hypergraph experts matching the scene’s motion and scale characteristics is sufficient to refine cross-modal features. Thus we adopt Top-2 sparse routing as our default strategy to strike a balance between representation capacity and computational overhead.

\paragraph{MoE Auxiliary Loss.}
\input{tab6}
Table \ref{tab:tab6} verifies the necessity of the two auxiliary loss terms for stabilizing FG-MoE training. Adding the load-balancing loss $\mathcal{L}_{bal}$ alone evenly distributes training samples across all experts, lifting mAP to 52.1. The router Z-loss $\mathcal{L}_{z}$ independently mitigates numerical instability caused by oversized router logits and softmax saturation, bringing mAP up to 52.0. Combining both regularization terms achieves the best 52.5 mAP, as they jointly constrain balanced expert activation and stable routing probability distribution, enabling end-to-end stable training of the sparse hypergraph MoE framework without expert collapse.

\subsection{Qualitative Results}
\begin{figure}[t]
    \centering\includegraphics[width = 1\linewidth]{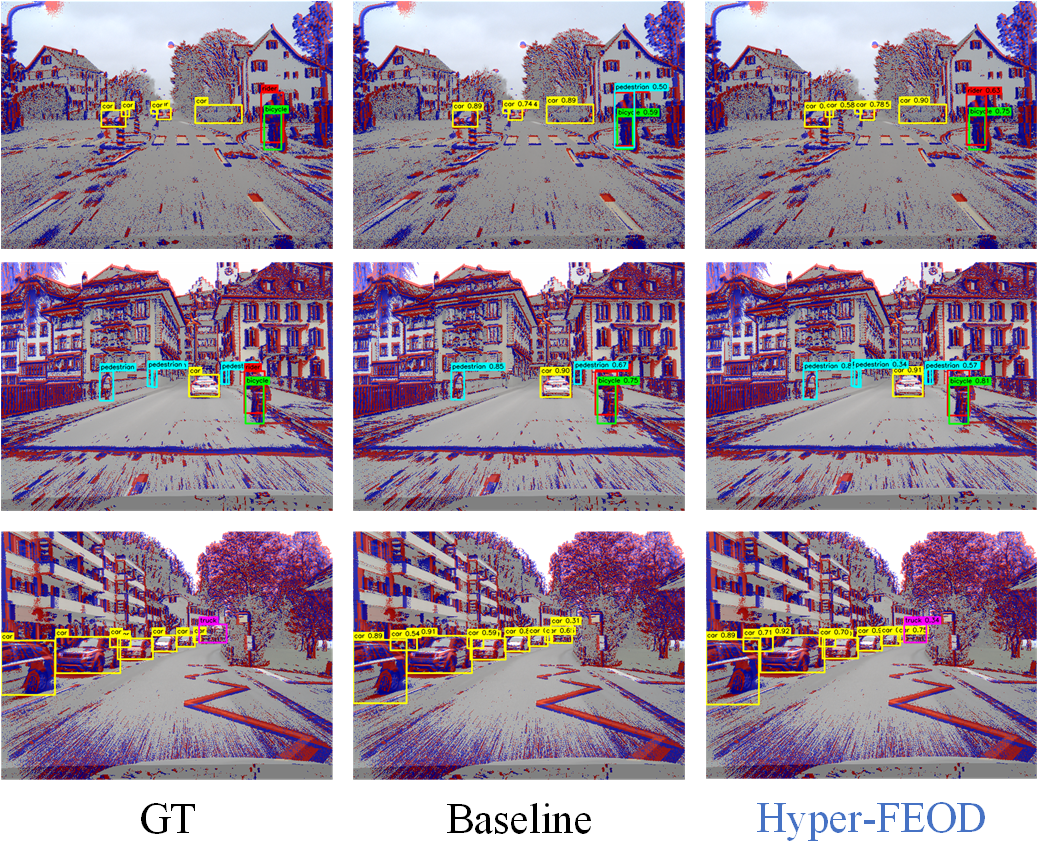}
    \caption{Visual comparison of detection results between the baseline and HEDFusion on the DSEC-Detection dataset.}
    \label{fig:vis}
    \vspace{-0.3cm}
\end{figure}
We provide visual detection comparisons on DSEC-Detection in Fig.~\ref{fig:vis} , covering low-light urban, dense building and complex background traffic scenes. The baseline method frequently misses small, weakly illuminated traffic participants and generates inaccurate loose bounding boxes for overlapping objects. In contrast, Hyper-FEOD accurately retains motion-relevant targets and suppresses false positive background responses. This qualitative observation aligns with our quantitative mAP improvement: the SHCF module concentrates high-order cross-modal modeling while FG-MoE handles dynamic scene.

\subsection{Limitations}
Despite its superior performance, Hyper-FEOD has a few limitations. First, the event-guided node sampling relies on active event streams; in scenarios with near-zero motion or severe event sparsity, adaptive target selection may degrade. Second, storing multiple expert weights increases the overall memory footprint, which may constrain deployment on edge devices. To address these challenges, future work will explore robust cross-modal target selection under event-sparse conditions, as well as lightweight MoE techniques.

%% file: tab1.tex
\renewcommand{\arraystretch}{1.3}
\begin{table*}[t]
\centering
\setlength{\tabcolsep}{2.5pt} 
\footnotesize 
    \caption{Comparison with state-of-the-art methods on PKU-DAVIS-SOD and DSEC-Detection. Runtime
    was measured with batch size 1 on an NVIDIA Tesla V100. For Hyper-FEOD, the parameter
    entry marked by $\dagger$ is reported as stored/active under Top-2
    inference; Best results
    are in bold.}
\vspace{-0.2cm}
\begin{tabular}{ccccccccc}
\toprule
\multirow{2}{*}{Method} & \multicolumn{3}{c}{PKU-DAVIS-SOD~\cite{li2023sodformer}} & \multicolumn{3}{c}{DSEC~\cite{gehrig2024low}} & \multirow{2}{*}{Params (M)} \\
\cmidrule(lr){2-4} \cmidrule(lr){5-7}
& mAP & mAP50 & Time (ms) & mAP & mAP50 & Time (ms) & \\
\midrule

JDF~\cite{li2019event} & - & 44.2 & - & - & - & - & $>60^*$ \\
FPN-Fusion~\cite{tomy2022fusing} & 29.0 & 57.5 & - & - & - & - & - \\
SFNet~\cite{liu2024enhancing} & 25.7 & 50.3 & - & 30.4 & 51.4 & - & - \\
EOLO~\cite{cao2024chasing} & 19.4 & 44.8 & - & - & - & - & - \\
NRE-Net~\cite{liu2025beyond} & 29.0 & 57.5 & - & - & - & - & - \\

SODFormer~\cite{li2023sodformer} & 20.7 & 50.4 & 39.7 & - & - & - & 82.5 \\
DAGr~\cite{gehrig2024low} & - & - & - & 37.6 & - & 16.8 & 19.0 \\
EFNet~\cite{sun2022event} & 26.4 & 52.9 & 13.7 & 30.1 & 47.7 & 16.5 & 33.0 \\
AFNet~\cite{zhang2023frame} & 27.5 & 53.3 & 14.5 & 31.4 & 48.8 & 19.7 & 60.1 \\
ReNet~\cite{zhou2022rgb} & 28.8 & 54.9 & 14.2 & 31.6 & 49.0 & 17.5 & 59.8 \\
FF-KDT~\cite{wang2024towards} & 28.5 & 54.2 & 18.0 & 33.6 & 52.5 & 19.6 & 38.0 \\
FAOD~\cite{zhang2024frequency}& 30.5 & 57.5 & \textbf{13.6} & 42.5 & 63.5 & \textbf{14.6} & 20.3 \\
\midrule
Hyper-FEOD (Ours) & \textbf{32.3} & \textbf{58.5} & 14.6 & \textbf{52.5} & \textbf{73.2} & 18.5 & \textbf{19.42\,/\,14.42}$^\dagger$ \\
\bottomrule
\end{tabular}
    \label{tab:tab1}
\vspace{-0.3cm}
\end{table*}

%% file: tab2.tex
\begin{table}[t]
    \centering
    \setlength{\tabcolsep}{8pt}
    \caption{Ablation study of Hyper-FEOD on DSEC-Detection. In Row 3, enabling only FG-MoE corresponds to using simple convolutional layers for multimodal fusion.}
    \label{tab:tab2}
    \begin{tabular}{ccccc}
        \toprule
        Row & SHCF & FG-MoE & mAP & mAP50 \\
        \midrule
        1 & &  & 50.9 & 71.5 \\
        2 &\cmark &  & 51.6 & 72.3 \\
        3 & & \cmark & 51.9 & 72.5 \\
        4 &\cmark & \cmark & 52.5 & 73.2 \\
        \bottomrule
    \end{tabular}
\end{table}

%% file: tab3.tex
\begin{table}[t]
    \centering
    \setlength{\tabcolsep}{5pt}
    \caption{Ablation study of SHCF on DSEC-Detection. In Row 2, enabling only Sparse Sampling corresponds to using simple linear layer to refine sparse tokens.}
    \label{tab:tab3}
    \begin{tabular}{ccccc}
        \toprule
        Row & Sparse Sampling & Hypergraph & mAP & mAP50 \\
        \midrule
        1 & &  & 50.9 & 71.5 \\
        2 &\cmark &  & 51.2 & 71.9 \\
        3 & & \cmark & 51.4 & 72.2 \\
        4 &\cmark & \cmark & 51.6 & 72.3 \\
        \bottomrule
    \end{tabular}
\end{table}

%% file: tab4.tex
\begin{table}[t]
    \centering
    \setlength{\tabcolsep}{6pt}
    \caption{Ablation study for expert numbers on DSEC-Detection.}
    \label{tab:tab4}
    \begin{tabular}{ccccc}
        \toprule
        Experts & Top-$k$ & Params (M) & mAP & mAP50 \\
        \midrule
        4 & 2 & 16.07 & 50.9 & 71.7 \\
        6 & 2 & 17.73 & 51.9 & 72.6 \\
        8 & 2 & 19.42 & 52.5 & 73.2 \\
        \bottomrule
    \end{tabular}
\end{table}

%% file: tab5.tex
\begin{table}[t]
    \centering
    \setlength{\tabcolsep}{6pt}
    \caption{Ablation study on Top-$K$ selection with 8 experts on
    DSEC-Detection.}
    \label{tab:tab5}
    \begin{tabular}{cccc}
        \toprule
        Top-$K$ & Expert sparsity & mAP & mAP50 \\
        \midrule
        2 & 75\% & 52.5 & 73.2 \\
        4 & 50\% & 52.2 & 73.0 \\
        6 & 25\% & 52.2 & 72.8 \\
        8 & 0\% & 52.0 & 72.8 \\
        \bottomrule
    \end{tabular}
\end{table}

%% file: tab6.tex
\begin{table}[t]
    \centering
    \setlength{\tabcolsep}{12pt}
    \caption{Ablation of the MoE auxiliary loss on DSEC-Detection.}
    \label{tab:tab6}
    \begin{tabular}{cccc}
        \toprule
        $\mathcal{L}_{\mathrm{bal}}$ & $\mathcal{L}_{z}$ & mAP & mAP50 \\
        \midrule
        &  & 51.6 & 71.5 \\
        \cmark &  & 52.1 & 73.0 \\
        & \cmark & 52.0 & 72.8 \\
        \cmark & \cmark & 52.5 & 73.2 \\
        \bottomrule
    \end{tabular}
\end{table}

%% file: 5_conclusion.tex
\section{Conclusion}
In this paper, we presented Hyper-FEOD, a novel frame-event object detection framework that addresses the challenges of complex multi-modal interaction and semantic heterogeneity between RGB frames and event streams. By incorporating hypergraph computation and mixture-of-experts mechanisms into cross-modal fusion, our approach achieves both efficient high-order relational reasoning and content-adaptive representation learning. Specifically, the Sparse Hypergraph-enhanced Cross-Modal Fusion (SHCF) module leverages event activity as a computational prior to select motion-critical target nodes, enabling sparse yet comprehensive high-order feature interaction while drastically reducing spatial redundancy. Furthermore, the Fine-Grained Mixture-of-Experts (FG-MoE) module dynamically routes sample representations to specialized hypergraph experts with distinct connectivity patterns and scales, accommodating diverse spatial regions and scene dynamics with minimal computational overhead. Extensive quantitative and qualitative evaluations on two benchmark datasets demonstrate that Hyper-FEOD consistently outperforms existing SOTA methods by a significant margin.

%% file: aaai2027.bib
@String(PAMI = {IEEE Transactions on Pattern Analysis and Machine Intelligence})

@String(CVPR= {Proceedings of the IEEE Conference on Computer Vision and Pattern Recognition})

@String(ICCV= {Proceedings of the IEEE International Conference on Computer Vision})

@String(ECCV= {Proceedings of the European Conference on Computer Vision})

@String(NIPS= {Proceedings of the Advances in Neural Information Processing Systems})

@String(ICME = {Proceedings of the IEEE International Conference on Multimedia and Expo})

@String(RAL= {IEEE Robotics and Automation Letters})

@String(ICRA= {Proceedings of the IEEE International Conference on Robotics and Automation})

@article{liu2020deep,
  title={Deep learning for generic object detection: A survey},
  author={Liu, Li and Ouyang, Wanli and Wang, Xiaogang and Fieguth, Paul and Chen, Jie and Liu, Xinwang and Pietik{\"a}inen, Matti},
  journal={International journal of computer vision},
  volume={128},
  number={2},
  pages={261--318},
  year={2020},
  publisher={Springer}
}

@article{oksuz2020imbalance,
  title={Imbalance problems in object detection: A review},
  author={Oksuz, Kemal and Cam, Baris Can and Kalkan, Sinan and Akbas, Emre},
  journal={IEEE transactions on pattern analysis and machine intelligence},
  volume={43},
  number={10},
  pages={3388--3415},
  year={2020},
  publisher={IEEE}
}

@article{perot2020learning,
  title={Learning to detect objects with a 1 megapixel event camera},
  author={Perot, Etienne and De Tournemire, Pierre and Nitti, Davide and Masci, Jonathan and Sironi, Amos},
  journal=NIPS,
  volume={33},
  pages={16639--16652},
  year={2020}
}

@inproceedings{gehrig2023recurrent,
  title={{Recurrent Vision Transformers for Object Detection with Event Cameras}},
  author={Gehrig, Mathias and Scaramuzza, Davide},
  booktitle=CVPR,
  pages={13884--13893},
  year={2023}
}

@article{gehrig2024low,
  title={{Low-Latency Automotive Vision with Event Cameras}},
  author={Gehrig, Daniel and Scaramuzza, Davide},
  journal={Nature},
  volume={629},
  number={8014},
  pages={1034--1040},
  year={2024},
  publisher={Nature Publishing Group UK London}
}

@article{zhang2024frequency,
  title={{Frequency-Adaptive Low-Latency Object Detection Using Events and Frames}},
  author={Zhang, Haitian and Wang, Xiangyuan and Xu, Chang and Wang, Xinya and Xu, Fang and Yu, Huai and Yu, Lei and Yang, Wen},
  journal={arXiv preprint arXiv:2412.04149},
  year={2024}
}

@article{li2023sodformer,
  title={{SODFormer: Streaming Object Detection with Transformer using Events and Frames}},
  author={Li, Dianze and Tian, Yonghong and Li, Jianing},
  journal=PAMI,
  volume={45},
  number={11},
  pages={14020--14037},
  year={2023},
  publisher={IEEE}
}

@inproceedings{han2023vision,
  title={{Vision HGNN: An Image is More than a Graph of Nodes}},
  author={Han, Yan and Wang, Peihao and Kundu, Souvik and Ding, Ying and Wang, Zhangyang},
  booktitle=ICCV,
  pages={19878--19888},
  year={2023}
}

@article{lei2025softhgnn,
  title={Softhgnn: Soft hypergraph neural networks for general visual recognition},
  author={Lei, Mengqi and Wu, Yihong and Li, Siqi and Zheng, Xinhu and Wang, Juan and Gao, Yue and Du, Shaoyi},
  journal={arXiv preprint arXiv:2505.15325},
  year={2025}
}

@inproceedings{fixelle2025hypergraph,
  title={Hypergraph Vision Transformers: Images are More than Nodes, More than Edges},
  author={Fixelle, Joshua},
  booktitle=CVPR,
  pages={9751--9761},
  year={2025}
}

@ARTICLE{hgnn,
  author={Gao, Yue and Feng, Yifan and Ji, Shuyi and Ji, Rongrong},
  journal=PAMI, 
  title={{HGNN$^+$: General Hypergraph Neural Networks}}, 
  year={2023},
  volume={45},
  number={3},
  pages={3181-3199}
}

@article{feng2024hyper,
  title={{Hyper-YOLO: When Visual Object Detection Meets Hypergraph Computation}},
  author={Feng, Yifan and Huang, Jiangang and Du, Shaoyi and Ying, Shihui and Yong, Jun-Hai and Li, Yipeng and Ding, Guiguang and Ji, Rongrong and Gao, Yue},
  journal=PAMI,
  year={2024},
  publisher={IEEE}
}

@article{lei2025yolov13,
  title={{YOLOv13: Real-Time Object Detection with Hypergraph-Enhanced Adaptive Visual Perception}},
  author={Lei, Mengqi and Li, Siqi and Wu, Yihong and Hu, Han and Zhou, You and Zheng, Xinhu and Ding, Guiguang and Du, Shaoyi and Wu, Zongze and Gao, Yue},
  journal={arXiv preprint arXiv:2506.17733},
  year={2025}
}

@inproceedings{tomy2022fusing,
  title={Fusing event-based and rgb camera for robust object detection in adverse conditions},
  author={Tomy, Abhishek and Paigwar, Anshul and Mann, Khushdeep S and Renzaglia, Alessandro and Laugier, Christian},
  booktitle={IEEE International Conference on Robotics and Automation},
  pages={933--939},
  year={2022},
  organization={IEEE}
}

@inproceedings{lin2014microsoft,
  title={{Microsoft COCO: Common Objects in Context}},
  author={Lin, Tsung-Yi and Maire, Michael and Belongie, Serge and Hays, James and Perona, Pietro and Ramanan, Deva and Doll{\'a}r, Piotr and Zitnick, C Lawrence},
  booktitle=ECCV,
  pages={740--755},
  year={2014},
  organization={Springer}
}

@inproceedings{li2019event,
  title={{Event-based Vision Enhanced: A Joint Detection Framework in Autonomous Driving}},
  author={Li, Jianing and Dong, Siwei and Yu, Zhaofei and Tian, Yonghong and Huang, Tiejun},
  booktitle=ICME,
  pages={1396--1401},
  year={2019},
  organization={IEEE}
}

@article{liu2024enhancing,
  title={{Enhancing Traffic Object Detection in Variable Illumination with Rgb-Event Fusion}},
  author={Liu, Zhanwen and Yang, Nan and Wang, Yang and Li, Yuke and Zhao, Xiangmo and Wang, Fei-Yue},
  journal=PAMI,
  year={2024},
  publisher={IEEE}
}

@inproceedings{cao2024chasing,
  title={{Chasing Day and Night: Towards Robust and Efficient All-Day Object Detection Guided by an Event Camera}},
  author={Cao, Jiahang and Zheng, Xu and Lyu, Yuanhuiyi and Wang, Jiaxu and Xu, Renjing and Wang, Lin},
  booktitle=ICRA,
  pages={9026--9032},
  year={2024},
  organization={IEEE}
}

@article{liu2025beyond,
  title={{Beyond RGB and Events: Enhancing Object Detection under Adverse Lighting with Monocular Normal Maps}},
  author={Liu, Mingjie and Liu, Hanqing and Zhu, Chuang},
  journal={arXiv preprint arXiv:2508.02127},
  year={2025}
}

@inproceedings{sun2022event,
  title={{Event-based Fusion for Motion Deblurring with Cross-modal Attention}},
  author={Sun, Lei and Sakaridis, Christos and Liang, Jingyun and Jiang, Qi and Yang, Kailun and Sun, Peng and Ye, Yaozu and Wang, Kaiwei and Gool, Luc Van},
  booktitle=ECCV,
  pages={412--428},
  year={2022},
  organization={Springer}
}

@inproceedings{zhang2023frame,
  title={{Frame-Event Alignment and Fusion Network for High Frame Rate Tracking}},
  author={Zhang, Jiqing and Wang, Yuanchen and Liu, Wenxi and Li, Meng and Bai, Jinpeng and Yin, Baocai and Yang, Xin},
  booktitle=CVPR,
  pages={9781--9790},
  year={2023}
}

@article{zhou2022rgb,
  title={{Rgb-Event Fusion for Moving Object Detection in Autonomous Driving}},
  author={Zhou, Zhuyun and Wu, Zongwei and Boutteau, R{\'e}mi and Yang, Fan and Demonceaux, C{\'e}dric and Ginhac, Dominique},
  journal=ICRA,
  year={2023}
}

@article{wang2024towards,
  title={{Towards Robust Keypoint Detection and Tracking: A Fusion Approach with Event-aligned Image Features}},
  author={Wang, Xiangyuan and Yu, Huai and Yu, Lei and Yang, Wen and Xia, Gui-Song},
  journal=RAL,
  year={2024},
  publisher={IEEE}
}

@article{gallego2022event,
  author  = {Guillermo Gallego and Tobi Delbruck and Garrick Orchard and Chiara Bartolozzi and Brian Taba and Andrea Censi and Stefan Leutenegger and Andrew J. Davison and J{\"o}rg Conradt and Kostas Daniilidis and Davide Scaramuzza},
  title   = {Event-Based Vision: A Survey},
  journal = {IEEE Transactions on Pattern Analysis and Machine Intelligence},
  volume  = {44},
  number  = {1},
  pages   = {154--180},
  year    = {2022},
  doi     = {10.1109/TPAMI.2020.3008413},
  url     = {https://arxiv.org/abs/1904.08405}
}

@article{gehrig2021dsec,
  author  = {Mathias Gehrig and Willem Aarents and Daniel Gehrig and Davide Scaramuzza},
  title   = {{DSEC}: A Stereo Event Camera Dataset for Driving Scenarios},
  journal = {IEEE Robotics and Automation Letters},
  volume  = {6},
  number  = {3},
  pages   = {4947--4954},
  year    = {2021},
  doi     = {10.1109/LRA.2021.3068942},
  url     = {https://dsec.ifi.uzh.ch/}
}

@article{cui2025peod,
  title={PEOD: A Pixel-Aligned Event-RGB Benchmark for Object Detection under Challenging Conditions},
  author={Cui, Luoping and Liu, Hanqing and Liu, Mingjie and Lin, Endian and Jiang, Donghong and Wang, Yuhao and Zhu, Chuang},
  journal={arXiv preprint arXiv:2511.08140},
  year={2025}
}

@article{shazeer2017outrageously,
  title={Outrageously large neural networks: The sparsely-gated mixture-of-experts layer},
  author={Shazeer, Noam and Mirhoseini, Azalia and Maziarz, Krzysztof and Davis, Andy and Le, Quoc and Hinton, Geoffrey and Dean, Jeff},
  journal={arXiv preprint arXiv:1701.06538},
  year={2017}
}

@article{fedus2022switch,
  title={Switch transformers: Scaling to trillion parameter models with simple and efficient sparsity},
  author={Fedus, William and Zoph, Barret and Shazeer, Noam},
  journal={Journal of Machine Learning Research},
  volume={23},
  number={120},
  pages={1--39},
  year={2022}
}

@article{riquelme2021scaling,
  title={Scaling vision with sparse mixture of experts},
  author={Riquelme, Carlos and Puigcerver, Joan and Mustafa, Basil and Neumann, Maxim and Jenatton, Rodolphe and Susano Pinto, Andr{\'e} and Keysers, Daniel and Houlsby, Neil},
  journal={Advances in Neural Information Processing Systems},
  volume={34},
  pages={8583--8595},
  year={2021}
}

@article{zoph2022st,
  title={St-moe: Designing stable and transferable sparse expert models},
  author={Zoph, Barret and Bello, Irwan and Kumar, Sameer and Du, Nan and Huang, Yanping and Dean, Jeff and Shazeer, Noam and Fedus, William},
  journal={arXiv preprint arXiv:2202.08906},
  year={2022}
}
